\documentclass[10pt,twocolumn,letterpaper]{article}

\usepackage{iccv}

\usepackage{times}
\usepackage{epsfig}
\usepackage{graphicx}
\usepackage{amsmath}
\usepackage{amssymb}
\usepackage{color}
\usepackage[lofdepth,lotdepth]{subfig}
\usepackage{multirow}
\usepackage{tabularx} 
\usepackage{booktabs}
\usepackage{enumitem}
\usepackage[super]{nth}
\usepackage[numbers,sort&compress]{natbib}

\definecolor{arthurcolor}{RGB}{0, 80, 160}
\definecolor{lchcolor}{RGB}{143, 188, 143}

\usepackage[dvipsnames]{xcolor}

\usepackage[pagebackref=true,breaklinks=true,letterpaper=true,colorlinks,bookmarks=false]{hyperref}

\iccvfinalcopy 

\def\iccvPaperID{****} 

\ificcvfinal\fi

\begin{document}

\title{Selective Feature Compression for Efficient Activity Recognition Inference}

\author{Chunhui Liu\thanks{Equally Contribute}, Xinyu Li\footnotemark[1], Hao Chen, Davide Modolo, Joseph Tighe\\
Amazon Web Services\\
{\tt\small \{chunhliu, xxnl, hxen, dmodolo, tighej\}@amazon.com}}
\maketitle

\begin{abstract}
Most action recognition solutions rely on dense sampling to precisely cover the informative temporal clip. Extensively searching temporal region is expensive for a real-world application. In this work, we focus on improving the inference efficiency of current action recognition backbones on trimmed videos, and illustrate that one action model can also cover then informative region by dropping non-informative features. We present Selective Feature Compression (SFC), an action recognition inference strategy that greatly increase model inference efficiency without any accuracy compromise. Differently from previous works that compress kernel sizes and decrease the channel dimension, we propose to compress feature flow at spatio-temporal dimension without changing any backbone parameters.
Our experiments on Kinetics-400, UCF101 and ActivityNet show that SFC is able to reduce inference speed by 6-7x and memory usage by 5-6x compared with the commonly used 30 crops dense sampling procedure, while also slightly improving Top1 Accuracy. We thoroughly quantitatively and qualitatively evaluate SFC and all its components and show how does SFC learn to attend to important video regions and to drop temporal features that are uninformative for the task of action recognition.

\end{abstract}

\section{Introduction}
\label{sec:intro}

\begin{figure}
    \centering
    \includegraphics[width=82mm]{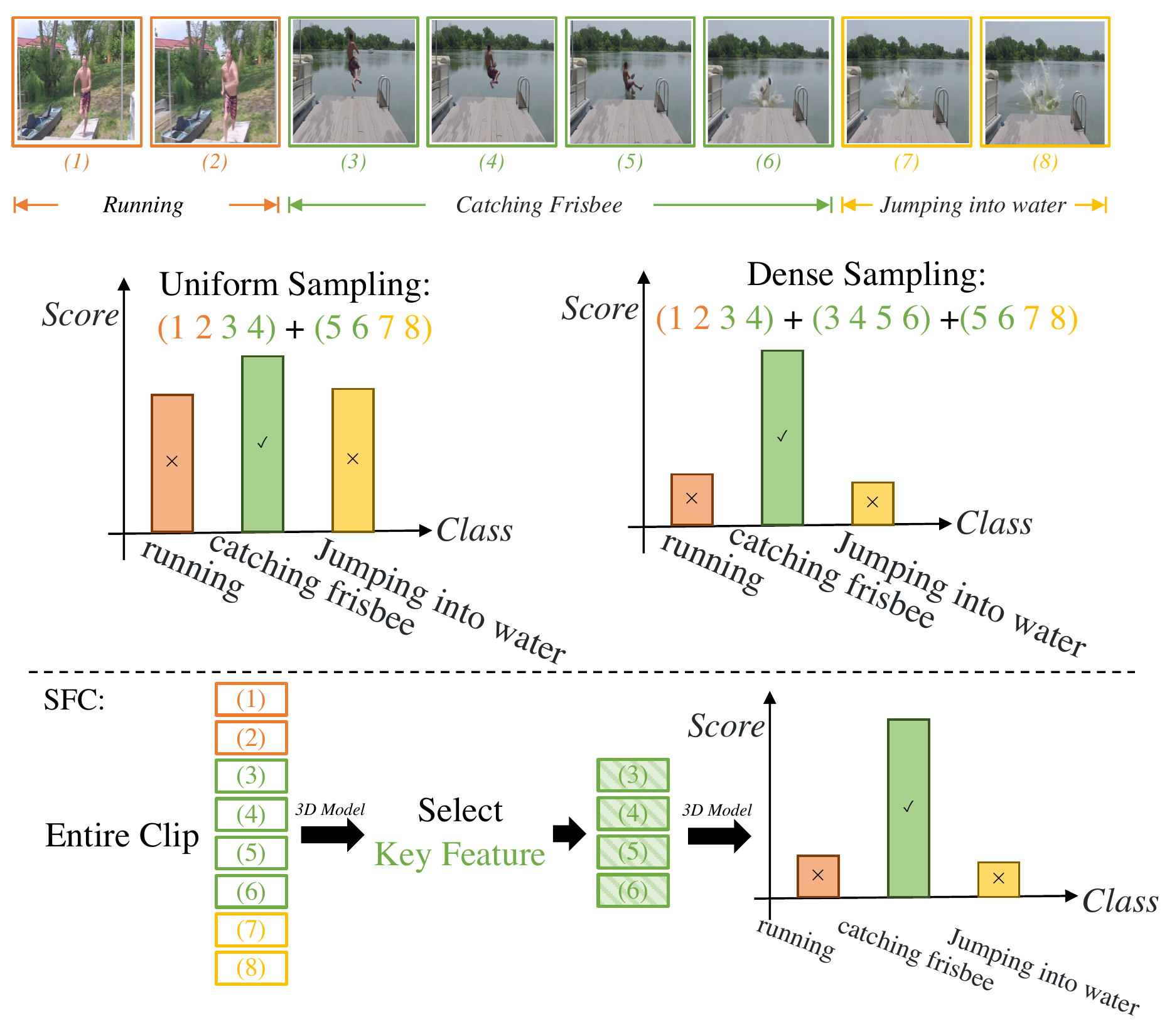}
    \caption{\it \small With fast motion, uniform sampling (row 1, left) is not guaranteed to precisely locate the key region. Consequently, an action model that uses it can be distracted by noise and can fail to recognize the correct action. Instead, dense sampling (row 1, right) looks at all the possible regions and can precisely locate the action and ignore noisy regions via voting. While accurate, dense sampling is however very inefficient. We propose SFC (row 2) to avoid sampling and instead it looks at the whole video and compresses its features into a representation that is smaller and adequate for action recognition.}
    \label{fig:teaser}
    \vspace{-5mm}
\end{figure}


Action recognition with 3D CNNs has seen significant advances in recent years~\cite{carreira2017quo, taylor2010convolutional,tran2018closer,xie1712rethinking,tran2015learning, qiu2017learning, luo2017thinet, wang2018non, feichtenhofer2018slowfast, wu2019multi, feichtenhofer2020x3d}, thanks to their ability to implicitly model motion information along with the semantic signals. 
However, 3D models require a substantial amount of GPU memory and therefore they cannot operate on a whole video sequence. Instead, they sample short video clips of 0.2-4 seconds and independently classify action labels for each clip. Finally, they pull the results from all the clips together and generate a final video-level prediction. How to sample these clips plays a critical role in these models and several techniques have been proposed (fig.~\ref{fig:teaser}, table~\ref{tab:method_cmp}). Among these,  \emph{dense sampling} is considered the best solution and employed in almost all action recognition models. Dense sampling works in a temporal sliding window manner and it over-samples overlapping segments exhaustively, ensuring that no information is missed and that the action is precisely localized within a window. While this achieves the best performance, it is also highly inefficient, as a lot of the sampled clips are highly related to each other (redundant) or not informative for the classification prediction of the video (irrelevant). 
Recently, SCSampler~\cite{korbar2019scsampler} and MARL~\cite{wu2019multi} proposed to learn to select a small subset of relevant clips to reduce this dense prediction problem. While they reach very promising results, their hard selection mechanism can potentially disregard useful information, especially on videos with complex actions. 

Differently, in this paper we propose a novel technique that reduces the dense sampling problem via feature compression. We call this {\it Selective Feature Compression} (SFC). SFC looks at long video sequences (up to two minutes) and compresses the rich information of their frames into much smaller representations, which are then analyzed by a more expensive 3D network that predicts the corresponding video action labels (fig.~\ref{fig:teaser}). This leads to a significant improvement in inference speed, without any drop in accuracy. We achieve this by inserting SFC within pre-trained action recognition models and by compressing their internal representations. In details, we propose to split an action network into two sub-networks (head and tail) and place  SFC  in  between  them. Then, FCP compresses the spatial-temporal feature  from  the  head  network  along  the temporal dimension using an attention-based mechanism that learns global feature correlation within the entire video sequence. Finally, this compressed representation is passed the tail network, which now operates on a much smaller input. 
This design brings the additional benefit  of not needing to re-train the already trained action network, as SFC can be finetuned independently and very quickly. Furthermore, SFC offers a good trade-off between inference and performance that can be easily tuned based on necessity (e.g., for fast real-world applications we can compress more aggressively at the cost of some performance).  


To validate the effectiveness of SFC, we present an extensive analysis on the popular Kinetics 400 dataset~\cite{carreira2017quo} and UCF101~\cite{ucf101} datasets. Our results show that SFC works with a wide range of backbones and different pre-trainings. SFC maintains the same top-1 accuracy of dense sampling (30 crops), while improving the inference throughput by 6-7$\times$ and reducing the memory usage by 6$\times$. While we designed SFC to replace the dense sampling strategy for action recognition on short trimmed videos, we also investigate its applicability on longer untrimmed content. We present results on ActivityNet~\cite{caba2015activitynet} datasets, where we show that SFC can be used in conjunction with uniform sampling and improve both performance and runtime. Finally, we present a visual analysis showing how SFC focuses on the informative parts of a video during feature compression.

\section{Related Work}\label{sec:rl}
As introduced in the previous section, sampling plays a critical role in action recognition (table~\ref{tab:method_cmp}).
Early video classification models were trained on individual frames, as opposed to video clips, and\textit{sparsely sampling frames}~\cite{karpathy2014large} uniformly or within segment-based clips~\cite{wang2016temporal} were popularly used. 
Although efficient, sparse frame sampling does not work well with 3D networks~\cite{carreira2017quo, taylor2010convolutional,tran2018closer,xie1712rethinking,tran2015learning, qiu2017learning}, as it breaks the temporally continuity required by these 3D models. 
The most intuitive way to run video inference with 3D networks is to take the entire video as input and run \textit{fully convolutional inference} \cite{Wang_2016_CVPR, yu2017fully, liu2017online, hu2017temporal}. However, the fully convolutional inference does not scale well for long videos as the memory requirements are well above the capabilities of modern GPUs~\cite{wang2018non, feichtenhofer2018slowfast}.
Inspired by image classification, multi-crop \textit{dense sampling} inference~\cite{wang2018non} improved accuracy and reduced memory need per iteration, by performing fully convolutional inference on a sampled set of video crops. While this procedure remains one of the most widely used~\cite{feichtenhofer2018slowfast,li2020directional, wang2018non, li2020tea, liu2020tam, yang2020temporal, feichtenhofer2020x3d}, the memory consumption and computation complexity in video level are instead greatly increased.

While the research on the topic of efficient inference remains limited, it is starting to gain more attention. One typical inference speed-up strategy is using \textit{kernel compression} to reduce the total computation required for one pass of the network~\cite{luo2017thinet,kim2018paraphrasing}. Some other methods reduce the computation by distilling the efficient convolution kernel~\cite{zolfaghari2018eco} or use 2D convolution for spatial-temporal modeling by temporal feature shifting~\cite{lin2019tsm, alizadeh2012less}. However, these special customized modules need fully re-training the backbone weights, making them rather difficult to generalize and to benefit from recent progress on model design and data. Differently from these methods, SFC does not require manipulating the backbone network weights, and it is much simpler. 
Moreover, note that kernel compression methods still employ the heavy multi-crop inference and they can benefit from this paper's new way of doing a single pass inference using SFC.

\begin{table}[t]
\centering
 \resizebox{\columnwidth}{!}{
        \begin{tabular}{lcccc}
        \toprule
          Method & Accuracy & Memory  & Latency & Generalization \\
        \midrule
        Sparse Sampling    & \color{YellowOrange} Medium      & \color{Green} Low         & \color{Green} Low     & \color{Green} High             \\
        Fully Convolutional & \color{YellowOrange} Medium      & \color{YellowOrange} Medium         & \color{Green} Low     & \color{Green} High             \\
        Dense Sampling     & \color{Green} High     & \color{BrickRed} High        & \color{BrickRed} High    & \color{Green} High             \\
        Kernel Compression & \color{YellowOrange} Medium   & \color{YellowOrange} Medium      & \color{YellowOrange} Medium  & \color{BrickRed} Low             \\
        Input Sampling     & \color{Green} High     & \color{YellowOrange} Medium      & \color{YellowOrange} Medium  & \color{Green} High             \\
        \midrule
        This paper: SFC     & \color{Green} High     & \color{Green} Low         & \color{Green} Low     & \color{Green} High        \\
        \bottomrule
        \end{tabular}}
    \caption{\it \small Comparison of different inference strategies. Our SFC module tries to improve action recognition inference along all these aspects.\vspace{-2mm}}
            \vspace{-2mm}
    \label{tab:method_cmp}
\end{table}



Following the idea of input sampling used for other video tasks~\cite{gong2014diverse,jain2014action,mahasseni2017unsupervised,zhang2016real}, another way to boost inference speed is through applying strategically \textit{input sampling} on the input video.
For example, \cite{wu2019multi} uses a light weight network to sample a few frames and \cite{korbar2019scsampler} uses it to sample sub-clips, achieving about $2\times$ speed-up and maintaining accuracy. 
However, those methods involve a two-stage training which includes learning a selection network by reinforcement learning or oracle loss design. The added selection network brings additional computation, which limits the efficiency and memory improvement. Our SFC shares the same core idea of improving inference by selecting relevant video information, but it does so at the feature level. 




\begin{figure*}[t]
    \centering
    \includegraphics[width=0.9\textwidth]{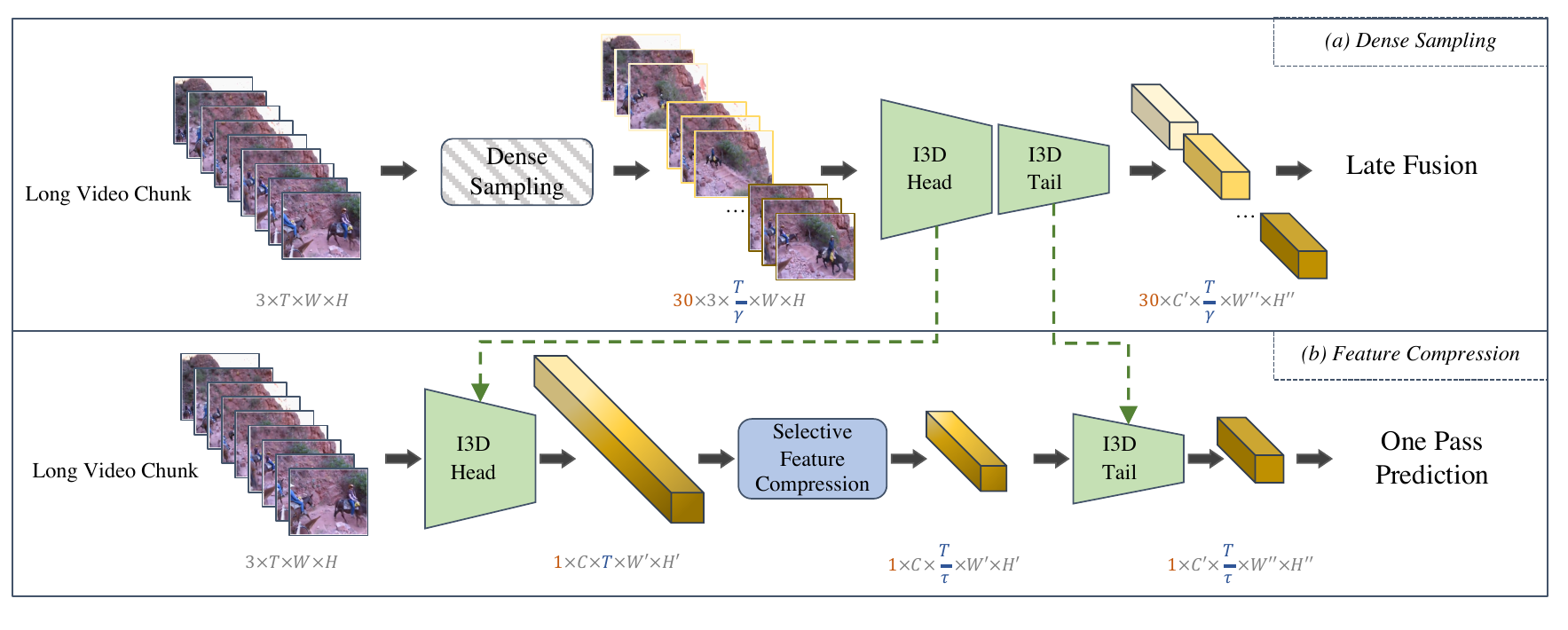}
     \caption{\it \small Dense sampling inference vs our SFC procedure. Instead of running inference on multiple crops and (late) fusing their predictions, SFC runs inference once on the whole video sequence. Moreover, the SFC is placed within the action recognition network, so that it can compress the output of I3D Head and reduce inference time of I3D Tail.}
     \label{fig:archi}
         \vspace{-3mm}
\end{figure*}

\section{Learn to look by feature compression}\label{sec:SFC}



As introduced before, some of the most popular and successful action recognition approaches~\cite{carreira2017quo, taylor2010convolutional,tran2018closer,xie1712rethinking,tran2015learning, qiu2017learning} train on short video crops that randomly sampled from long training videos. At inference, they densely sample a test video into a number of crops (i.e., 30~\cite{wang2018non,feichtenhofer2018slowfast,feichtenhofer2020x3d} and 10~\cite{korbar2019scsampler} for Kinetics), run inference on each of these independently, and finally average their predictions into a video-level output (Fig.~\ref{fig:archi}{\it \small top}). 
While these methods have successfully achieve outstanding action recognition performance, they lack in inference efficiency. Instead, we propose an efficient solution that removes the need for dense sampling, while also reducing the inference time of the 3D backbone by compressing its features (fig.~\ref{fig:archi}{\it \small bottom}).

%


We achieve this by splitting a pre-trained 3D network encoder into two components, {\it head} and {\it tail}, and inserting a novel Selective Feature Compression module in between them. Our intuition for SFC is twofold: (i) we can remove redundant information and preserve the useful parts directly at the feature level and (ii) by compressing the features we can reduce the inference computation of {\it tail} considerably, as it can now operate on a much smaller input. Formally, given an input video $V$, our approach predicts activity labels $a$ as follows:
\begin{equation}
    a = \Theta_{tail}(\Phi(\Theta_{head}(V))),
\end{equation} 
where $\Theta_{head}$ and $\Theta_{tail}$ represent the decoupled head and tail components and $\Phi$ represents our SFC.
And the feature selection operation is learnt by a cross entropy loss using original action labels $Y$:
\begin{equation}
    \operatorname{min}_{\Phi} \;\; \mathcal{L_{\operatorname{CE}}}\left(Y; \Theta_{tail}(\Phi(\Theta_{head}(V)) \right)
\end{equation} 

Our design has two additional advantages: (1) it achieves fast inference by encoding the entire video clip in a single pass, as opposed to splitting it into crops and processing each of them independently; (2) it does not require re-training of the 3D backbone thus is flexible to use. In the following part of this section we 
present our proposed highly effective, attention-based design (Sec.~\ref{sec:3.3}).

\begin{figure}[t]
    \centering
    \includegraphics[width=0.48\textwidth]{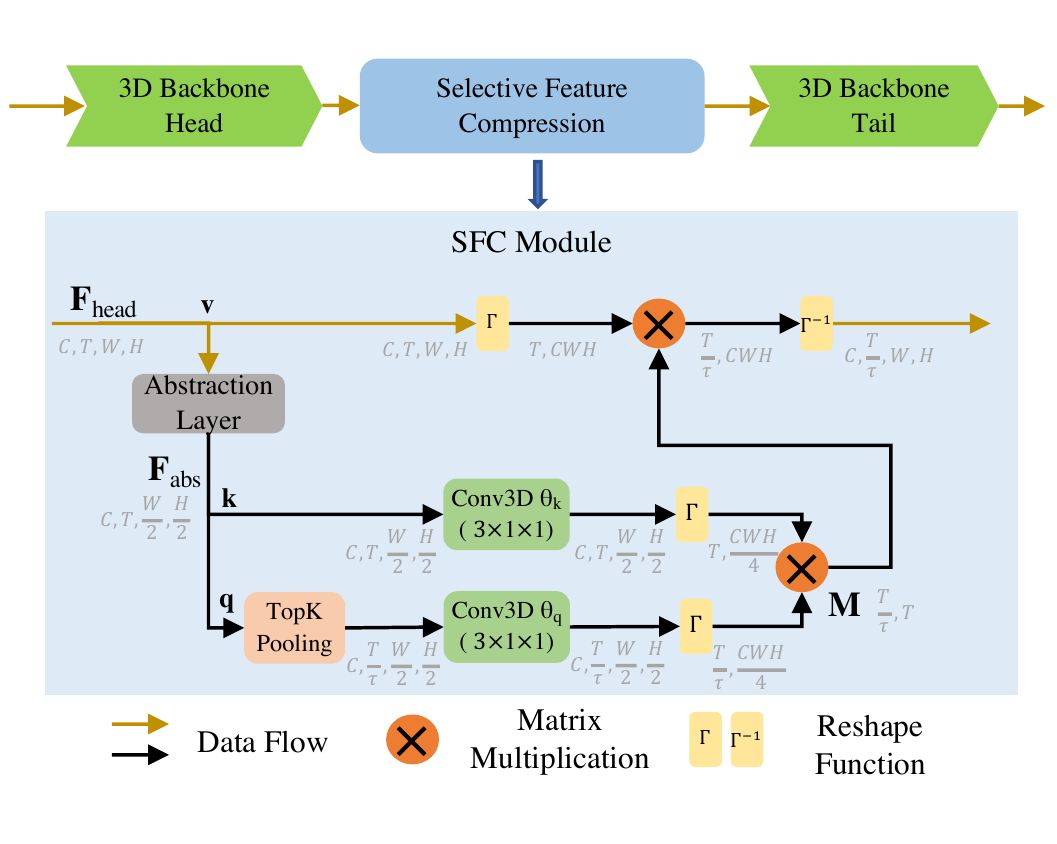}
    \caption{\it \small Selective Feature Compression design. Our proposed design is capable of modelling long-distance interactions among features and better discover the most informative video regions.}
        \label{fig:sfc}
\end{figure}

\subsection{Selective Feature Compression}
\label{sec:3.3}


Our solution is inspired by the self-attention mechanism~\cite{vaswani2017attention, wang2018non}, as it can directly model long-distance interactions between features to automatically discover what to attend to, or in our case, what to compress. However, differently from the original formulation of self-attention that takes an input sequence of length $n$ and outputs the sequence in the same length \cite{wang2018non}, we want SFC to return a 
shorter (compressed) sequence of length $n/\tau$. Furthermore, we also want SFC to output compressed features that are compatible with the frozen {\it tail}. 
To satisfy both requirements, we model SFC as shown in Fig.~\ref{fig:sfc}. 
Formally, given the feature generated by the {\it head} sub-network  $\mathbf{F}_{head} = \Theta_{head}(V) \in \mathbb{R}^{C \times T \times W \times H}$, we formulate SFC as:
\begin{flalign}
    \Phi &= \operatorname{SFC}(\mathbf{k}=\mathbf{q}=\mathbf{F}_{abs}, \mathbf{v}=\mathbf{F}_{head}) \\
    &= \mathbf{M} \cdot \mathbf{v}, \label{eq:SFC} \\ 
    \mathbf{M} &= \operatorname{softmax}( \theta_{q}(\operatorname{pool}(\mathbf{q}))^T \cdot  \theta_{k}(\mathbf{k}) ), \label{eq:M}
\end{flalign} where $\theta_k$ and $\theta_q$ are linear transformations implemented as $(3\times1 \times1)$ 3D convolution kernels, $\operatorname{pool}$ refers to our TopK Pooling module, $\mathbf{F}_{abs}=\Theta_{abs}(\mathbf{F}_{head})$ are the head features re-encoded with our abstraction network and $\mathbf{M}$ is the attention map (note: we omit the reshape function $\Gamma$ in the equation for simplicity, but include it in Fig.~\ref{fig:sfc}).

Although this looks similar to a classic self-attention/non-local block, as $\mathbf{A}(\mathbf{k}=\mathbf{q}=\mathbf{v}=\mathbf{F}_{head}) = \mathbf{M} \cdot \theta_v(\mathbf{v})$, with $\mathbf{M} = \operatorname{softmax}( \theta_{q}(\mathbf{q})^T \cdot \theta_{k}(\mathbf{k}))$, our SFC is designed for a different purpose, and differs in the following key ways that are critical to our methods effectiveness.
We also present ablation studies in Section \ref{sec:res} to show that those details are crucial for a high performance. \\

\noindent{\bf Abstraction layer $\Theta_{abs}$.}  $\Theta_{head}$ is trained for action classification and is therefore natural that $\mathbf{F}_{head}$ (its output) captures information that is important for that task, like motion patters and object/place semantic. However, we argue that this information is not optimal for feature compression and to help our SFC module to generate a better attention map $\mathbf{M}$, we introduce an abstraction module that re-encodes $\mathbf{q}$ and $\mathbf{k}$ using two ResNet Blocks. This transforms the features from low-level motion patterns to more meaningful representations for compression ($\mathbf{F}_{abs}$). Finally, note that we only re-encode $\mathbf{k}$ and $\mathbf{q}$ using this layer, as we want to (i) specialize it for compression and (ii) preserve the features of $\mathbf{v}$ for compatibility with $\Theta_{tail}$. \\

\noindent{\bf TopK Pooling.} We use pooling to downsample the features of the query $\mathbf{q}$ from $T$ to $T/\tau$ in the temporal dimension. This ensures that the output of SFC is a compressed vector.
Instead of using average/max pooling which compress locally within a small window, we propose to use TopK pooling to allow the compression to select features that are temporally consecutive. 
Given the feature $\mathbf{F}_{abs} \in \mathbb{R}^{C \times T  \times W  \times H}$, TopK pooling (with downsample ratio $(\tau,1,1)$) returns $\mathbf{F}_{pool} \in \mathbb{R}^{C\times \frac{T}{\tau}\times W\times H}$ that contains the top $T/\tau$ highest activated features along time $T$.  \\


\noindent{\bf Value $\mathbf{v}$.} In addition to not re-encoding $\mathbf{v}$ using $\Theta_{abs}$, we also avoid transforming $\mathbf{v}$ with $\theta_{v}$, as this could also break compatibility with the $\Theta_{head}$. Instead, we train SFC to directly attend to $\mathbf{v}$ using the attention map $\mathbf{M}$ (eq.~\ref{eq:SFC}).


\section{Experimental Settings}
\noindent \textbf{Backbones.} To test the generalization ability of SFC, we plug it into some of the most popular activity recognition backbones: Slow-Only I3D-50 \cite{feichtenhofer2018slowfast}, TPN-50~\cite{yang2020temporal}, and R(2+1)D-152~\cite{tran2018closer}. For Slow-Only I3D, we experiment with three sample rates: 2, 4, and 8 with input length equal to 32, 16, and 8 respectively. For Slow-Only I3D and TPN, we use the publicly pre-trained models released in \cite{yang2020temporal}, where Slow-Only 16$\times$4 works slightly better than 32$\times$2. For R(2+1)D-152, we use the model pre-trained on IG-65M and released in \cite{ghadiyaram2019large}. 

\noindent \textit{Baseline.} 
For dense sampling, we follow the literature~\cite{wang2018non,feichtenhofer2018slowfast} that uniformly crops 10 clips temporally and 3 clips spatially, for a total of 30 crops. Please note that numbers for Slow-Only network is slightly different from \cite{feichtenhofer2018slowfast}. This is caused by slight variations in the videos of the Kinetics dataset, as some have been removed from YouTube. \\

\noindent \textbf{Implementation details.}
We start with a pre-trained 3D backbone and divide it into {\it head} and {\it tail} (using residual block 3 as the cutting point). We freeze head and tail weights (including BN) and insert our SFC between them. 

During \textbf{\textit{training}}, we generate gradients for SFC and {\it tail}, but only update the weights of SFC. We use the same augmentation and hyper-parameters that is used to train the original backbone, but pass the whole video as input instead of short clips. For training, we use 32 Tesla V100 GPUs for 15 epochs only, as SFC is light-weight and it converges fast. We set the initial learning rate to 0.01
and drop it by a factor of 10 at epoch 8 and again at epoch 12. We use an SGD optimizer with the weight decay set to 1e-5 and momentum to 0.9. To avoid over-fitting, we apply several data augmentations, like outward scaling, cropping ($224\times273$) and horizontal flipping. During \textbf{\textit{inference}}, instead, we resize each video to $256\times312$. \\

\begin{figure}[t]
\centering
    \includegraphics[width=.44\textwidth]{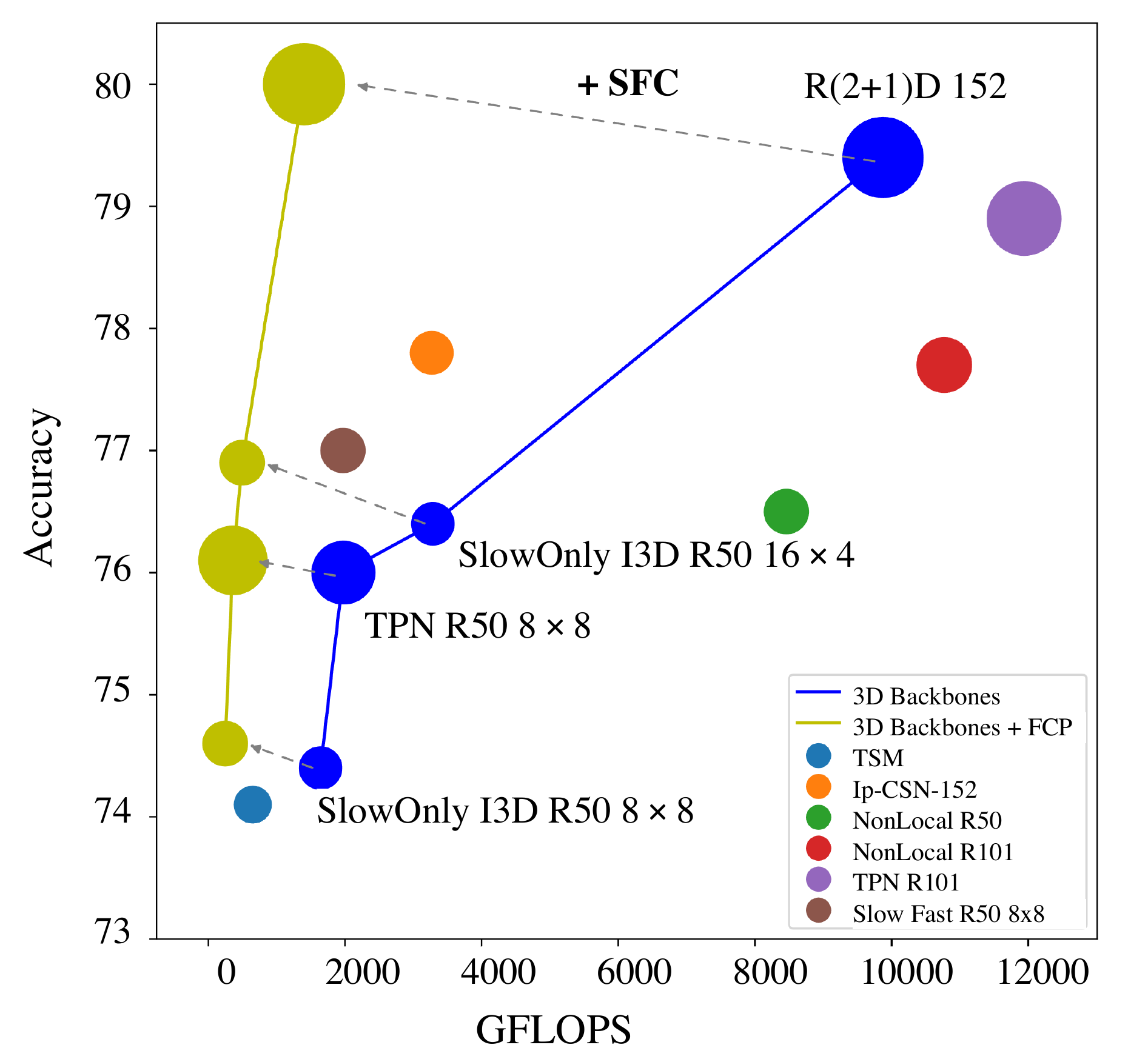}
    \caption{\it \small Inference speed (GFLOPS) vs Performance (Top1 Accuracy) vs Model size (Number of parameters, bubble size). Solid lines link the backbones evaluated in table~\ref{tab:table1} and dashed arrows show the performance improvement between using 30 crops dense sampling (blue) and SFC (yellow) with those backbones.}
    \label{fig:bubble1}
    \vspace{3mm}
    \centering
        \includegraphics[width=.43\textwidth]{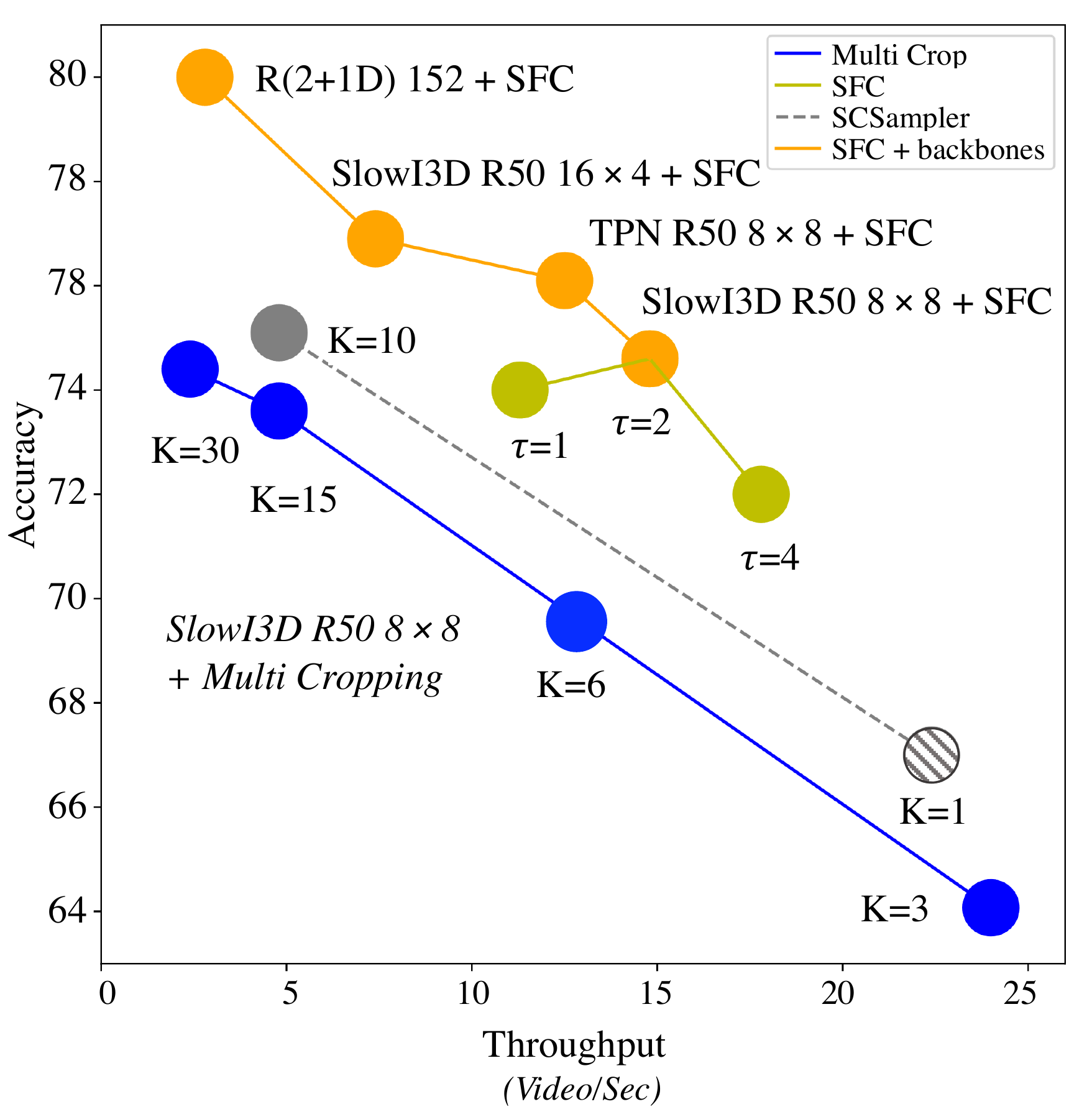}
    \caption{\it \small Comparing speed-accuracy trade-off with different sampling methods. We can observe that our proposed SFC provides a better inference solution.}
        \label{fig:bubble2}
    \vspace{-10mm}
\end{figure}

\noindent \textbf{Datasets.} We present results on three popular action recognition datasets: Kinetics 400~\cite{carreira2017quo}, UCF101~\cite{ucf101} and ActivityNet~\cite{caba2015activitynet}. Kinetics 400 consists of approximately 240k training and 20k validation videos trimmed to 10 seconds and annotated with 400 human action categories. UCF101 is a smaller dataset with 13k videos annotated with 101 action categories.
ActivityNet (v1.3)~\cite{caba2015activitynet} is an untrimmed dataset consisting of 19k videos, many of which are long (5 to 20 minutes), with 200 action classes. We report numbers on validation, as testing labels are not publicly available.   \\

\begin{table*}[t]
\centering
    \small
        \begin{tabular}{l|c|cc|rc|rc|cc}
            \toprule
             \multirow{2}*{Backbone} & \multirow{2}*{Trained} & \multirow{2}*{Inference} & \multirow{2}*{Input} & \multicolumn{2}{c|}{Efficiency} & \multicolumn{2}{c|}{Memory} & \multicolumn{2}{c}{Accuracy}\\
             &&&&FLOPS & Throughput & \#Params. & \#Videos & Top1 & Top5 \\
            \hline
            \multirow{2}*{Slow Only I3D-50 8$\times$8} & \multirow{2}*{K400} & 30 crops & 64$\times$30 & 1643G & 2.4 & 32.5M & 3  & 74.4&	91.4 \\
            & & SFC  & 288$\times$1  & \textbf{247G} & \textbf{14.8 (6$\times$)} & 35.9M & \textbf{19} & \textbf{74.6} & \textbf{91.4} \\
            \hline
            \multirow{2}*{Slow Only I3D-50 16$\times$4} & \multirow{2}*{K400} &  30 crops & 64$\times$30 & 3285G & 1.2 & 32.5M & 1 & 76.4	& 92.3  \\
            &  & SFC  &  288$\times$1 &\textbf{494G} & \textbf{7.4  (6$\times$)} & 35.9M  & \textbf{8} & \textbf{76.9}  & \textbf{92.5}\\
            \hline
                \multirow{2}*{Slow Only I3D-50 32$\times$2} & \multirow{2}*{K400} &  30 crops &  64$\times$30 & 6570G & 0.6 & 32.4M &  $<$1 &  75.7 & \textbf{92.3}  \\
            &  & SFC  & 288$\times$1 & \textbf{988G} & \textbf{3.6  (6$\times$)}&  35.9M  & \textbf{4}& \textbf{75.8} & 92.1 \\
            \hline
                \multirow{2}*{TPN-50 8$\times$8} & \multirow{2}*{K400} &  30 crops &  64$\times$30 & 1977G & 2.1 & 71.8M & 2.7 &  76.0 & \textbf{92.2}  \\
            &  & SFC  & 288$\times$1 & \textbf{359G} & \textbf{12.5 (6$\times$)}&  85.7M  & \textbf{17}& \textbf{76.1} & 92.1 \\
            \hline
                \multirow{2}*{R(2+1)D-152} & IG-65M$\rightarrow$ &  30 crops &  64$\times$30& 9874G & 0.4 &118.2M & $<$1 & 79.4 & 94.1\\
            &  K400 & SFC  & 288$\times$1& \textbf{1403G} & \textbf{2.8  (7$\times$)} & 121.7M & \textbf{5} & \textbf{80.0} & \textbf{94.5} \\
            \bottomrule
        \end{tabular}
        \vspace{-2mm}
        \caption{\it \small Comparison of different backbones with dense sampling inference and with SFC on Kinetics 400. Differently from the other entries, R(2+1) was initially pre-trained on the IG-65M datasets and later trained on K400. Given any backbone, we freeze it an train our SFC module only on K400 dataset. Finally, we compare the results with dense sampling (30 crops).}
        \label{tab:table1}
            \vspace{-3mm}
\end{table*}
\begin{table}
\small
    \centering
    \begin{tabular}{cccccc} 
			\toprule
		 $\tau$ & Data Used & FLOPs & Throughput & Top1 & Top5 \\ 
			\hline
			$1$ & 100\% & 352G & 9.6 & 73.8	&91.2 \\
			$4/3$ & 75\%  &299G & 12.1 & 74.5 & 91.4 \\
			\underline{$2$} & 50\%  &247G & 14.8 & \textbf{74.6} & \textbf{91.4} \\
			$4$& 25\% &195G & 17.8 & 72.0 & 90.1\\
			\bottomrule
		\end{tabular}
    \vspace{-2mm}
    \caption{\it \small The impact of different data using ratio to speed-accuracy tradeoff, using I3D-50 8$\times$8.}
    \label{tab:compression}
\end{table}
    
\begin{table}
\centering
\small
\begin{tabular}{lcr}
        \toprule
        Method & Throughput & Top1  \\ \toprule
        Baseline: 30 crops & 1$\times$ & 74.5 \\
        Single Crop & 30$\times$ & $-7.2$ \\
        \hline
        TSN Single Sampling \cite{wang2016temporal} & 30$\times$ & $-5.7$ \\
        TSN Dense Sampling \cite{wang2016temporal} & 3$\times$ & $-4.9$ \\
        SCSampler \cite{korbar2019scsampler} & 2$\times$ & $+2.5$ \\
        \hline
        \bf SFC & 6.2$\times$ & $+0.2$\\
        \bottomrule
        \end{tabular}
        \vspace{-2mm}
        \caption{\it \small Comparison with input sampling methods.}
        \label{tab:sampling}
\end{table}

\begin{table}[t]
    \centering
    \small
    \begin{tabular}{lccc}
    \toprule
        Method & FLOPS & \#Params. & Top1 \\
    \midrule
        TSM \cite{lin2019tsm}  & 64G $\times$ 10  & 24M & 95.9 \\
        I3D \cite{carreira2017quo}  & 65G $\times$ 30  & 44M & 92.9  \\
        NonLocal R50 \cite{wang2018non}  & 65G $\times$ 30  & 62M & 94.6  \\
    \midrule
        Slow I3D 8$\times$8, 1 Crop & 54G $\times$1 & 32.5M & 93.8  \\
        Slow I3D 8$\times$8, 30 Crops     & 54G $\times$30 & 32.5M & \textbf{94.5} \\
        Slow I3D 8$\times$8 + {\bf SFC} & 247G & 35.9M & \textbf{94.5} \\
    \bottomrule
    \end{tabular}
    \vspace{-2mm}
    \caption{\it \small Results on UCF 101. \vspace{-4mm}}
    \label{tab:ucf101}
\end{table}

\noindent \textbf{Evaluation Metrics.} In order to fully evaluate SFC for a practical usage, we use the following metrics:
\begin{enumerate}[itemsep=0pt,parsep=0pt,topsep=0pt, partopsep=0pt]
    \item {\it Accuracy.} We report Top1 and Top5 classification accuracy to evaluate action recognition performance. 
    \item {\it FLOPS.} We report floating point operations per-second to evaluate inference runtime. Note that we compute this over a video, while previous works reported it over clips. For fair comparison, we multiply clip-level FLOPS by $N$ when $N$ crops are sampled in dense sampling. 
    \item {\it Video Throughput}. As FLOPS cannot always estimate precise GPU speed due to different implementations of the model's operations, we also report video throughput speed, which is the number of videos that a single Tesla V100 GPU can process per second (excluding data loading).
    \item {\it Number of model parameters.} We use this number to report the complexity of a model.
    \item {\it Number of videos per batch}. In addition to model complexity, we also report the number of videos we can perform inference simultaneously (i.e., in a batch) on a single GPU (of 16GB, Tesla V100). This number show another perspective on the efficiency of an inference strategy. 
\end{enumerate}

\section{Experimental Results} \label{sec:res}
In this section we present a series of experiments. First, we show that SFC can considerably improve inference efficiency without affecting model accuracy (sec.~\ref{sec:k400efficient}) on two trimmed dataset: Kinetics-400 and UCF101. Then, we ablate the components of SFC to validate our design hypothesis (sec.~\ref{sec:k400ablation}) on Kinetics-400. 

\subsection{Results on Kinetics-400 and UCF101} \label{sec:k400efficient}

\noindent {\it Evaluating different backbones.}
We equip several popular action recognition backbones with SFC and evaluate their accuracy, inference efficiency and memory usage compared on Kinetics-400. We then compare their results to those obtained by the popular 30 crops dense sampling (table~\ref{tab:table1} figure ~\ref{fig:bubble1}). Results show that SFC generalizes well to different backbones, with different sampling rates (rows 1-3, using $8\times8$, $16\times4$ and $32\times2$), different backbone designs (rows 1, 4, 5, as SlowOnly, TPN, R(2+1)D) and different data pre-training (row 5, as uing IG-65M \cite{ghadiyaram2019large}). In general we observe that SFC improves the video-level inference efficiency by $6-7\times$ of dense sampling, without losing any performance. We argue that this is thanks to SFC 's ability to drop redundant and irrelevant information. Moreover, note how our compression design also reduces memory usage greatly, by around 7$\times$. This is because SFC does not need to analyze overlapping temporal windows and it can drop non-informative temporal regions. 

Results also show that a very expensive (and well-performing) R(2+1)D-152 solution can be run at the same throughput of the much cheaper I3D-50 8$\times$8, when equipped with SFC. In practice, thanks to SFC we are able to achieve the best accuracy at the same speed of a relatively fast model. In comparisos, Dense Sampling on R(2+1)D-152 requires 22GB of memory for a single video (assuming 30 crops), which is extremely expensive compared to the 3GB needed with SFC. \\


\noindent {\it Understanding speed-performance trade-off.}
One important factor in feature selection is about how the performance changes as we compress more and more. To understand this, we experiment with different compression ratio in table~\ref{tab:compression} and fig.~\ref{fig:bubble2}. Results show that some compression is actually beneficial and leads to small improvements in performance, as SFC is able to disregard useless information for classification. This was also observed in previous works~\cite{korbar2019scsampler, wu2019multi}. On the opposite end of the spectrum, compressing too much (i.e., $\tau$ = 4) leads to fast inference, but slightly lower performance. Compressing about 50\% of that data (i.e., $\tau$ = 2), offers a good trade-off and achieves the best overall performance on K400. \\

\noindent {\it Comparing with data selection.}
Here we compare our feature compression idea against data selection/sampling methods using I3D-50 8$\times$8 (table~\ref{tab:sampling}). We compare against TSN~\cite{wang2016temporal} sampling, which proposes to temporal segment the videos and extract discontinuous frames for 3D models, and the recent SCSampler, which uses a cheap network to select what segments to run full inference on. While TSN sampling is able to improve throughput, it also degrades performance quite severally, compared to the baseline. On the other hand, SCSampler and SFC can improve both the performance and the throughput. As SCSampler performs selection rather than compression, it disregard regions more aggressively than SFC and can achieve better performance than SFC, thought SFC is 3$\times$ more efficient. Finally, we would like to note that, in theory, selection methods (like SCSampler) work well on videos with simple actions, but may not generalize as well on more complex videos with a lot of important information. This because they are forced to (hard) select a subset of segments. On the other hand, SFC compresses by looking at the whole video, which can preserve all the rich information. \\


\noindent {\it Evaluating on UCF101.} Finally, we present results on UCF101 in table~\ref{tab:ucf101}. Similarly to the Kinetics-400 results, SFC is able to match the performance of dense sampling (30 crops), while being $6.5\times$ more efficient.  Overall, we believe that SFC offers a good trade-off between fast inference and competitive performance, which is one of the keys to enable action recognition for real-world applications.

\begin{table}[t]
	\centering
	\small
	 \subfloat[Backbone Split Choices]{
	 \small
    \begin{tabularx}{0.48\textwidth}{lcccc} 
			\toprule
			SFC Insertion Point & FLOPs & Throughput & Top 1& Top5  \\ 
			\hline
			Res 1234 / Res 5 & 305G & 14.0 & 73.7 & 90.7 \\
			\underline{Res 123 / Res 45} & 247G & 14.8 & \textbf{74.6} & \textbf{91.4} \\
			Res 12 / Res 345  & 211G & 15.2 & 73.2 & 90.5 \\
			Res 1 / Res 2345 & 163G & 21.2 & 73.1 & 90.9\\
			\bottomrule
		\end{tabularx}
		\label{tab:aba-a}
	} \hfill
	\subfloat[Pooling Strategy]{
	\small
	\begin{tabularx}{0.2\textwidth}{lc} 
			\toprule
			$\operatorname{pool}(\mathbf{q})$  & Top1   \\ 
			\hline
		 Average Pooling & 72.9    \\ 
		 Max Pooling &  72.6   \\ 
			\underline{TopK Pooling} & \textbf{74.6} \\
			\bottomrule
		\end{tabularx}
		\label{tab:aba-c}
	}
	\subfloat[KQV Choices]{
	\begin{tabularx}{0.27\textwidth}{lc} 
		\toprule
			Model   & Top1   \\ 
			\hline
			 $\mathbf{k}=\mathbf{q}=\mathbf{v}=\mathbf{F}_{bead}$ &   72.7  \\ 
			$\mathbf{k}=\mathbf{q}=\mathbf{v}=\mathbf{F}_{abs}$ & 63.2     \\ 
			\underline{$\mathbf{k}=\mathbf{q}=\mathbf{F}_{abs}, \mathbf{v}=\mathbf{F}_{head}$} & \textbf{74.6}    \\ 
			\bottomrule
	\end{tabularx}
	\label{tab:aba-d}
	} \hfill
	\subfloat[Different Conv Kernels]{
	\begin{tabularx}{0.355\textwidth}{lccc} 
		\toprule
			Model   &  FLOPS & \#Params. &Top1   \\ 
			\hline
			Conv (3x3x3) & 356G & 48.52M & 72.0  \\ 
			\underline{Conv (3x1x1)} &  247G & 35.93M &\textbf{74.6}  \\ 
			 Conv (1x1x1) & 238G & 34.89M & 72.5 \\ 
			\bottomrule
	\end{tabularx}
	\label{tab:aba-e}
	}
	\label{tab:ablation}
	\vspace{-2mm}
	\caption{\it \small Ablation studies on Kinetics-400, using I3D-R50 8$\times$8. \vspace{-5mm}}
\end{table}

\subsection{Ablation Study of SFC} \label{sec:k400ablation}
In this section we investigate how SFC performance and efficiency change with respect to some of its parameters. Specifically, we present an ablation study on 4 important components using a I3D-50 8$\times$8 backbone:\\

\noindent {\it Head/Tail Splitting point.} (table~\ref{tab:aba-a}). We investigate different splits of the action recognition backbone. Each option corresponds to a different locations for our feature compression plugin. Results show that all entries offer competitive results, with the best accuracy achieved by splitting after layer 3. Splitting after layer 1 offers the fastest throughput, at the cost of 1.5 Top1 accuracy. This is an interesting result for applications that require very fast inference. \\

\noindent {\it Pooling Strategy} (table~\ref{tab:aba-c}). We chose TopK pooling in SFC to temporally downsample the features of the query $\mathbf{q}$, because it can choose continuous temporal information and achieve the highest compatibility with \textit{tail}. Results show that Topk Pooling is indeed the best among the reported pooling strategies. \\

\noindent {\it Abstraction Layer and KQV choices.} (table~\ref{tab:aba-d}). SFC uses an abstraction layer to improve the features of the query $\mathbf{q}$ and the key $\mathbf{k}$ for compression. The features of $\mathbf{v}$ remain instead unchanged, as if they were to be transformed, they would lose compatibility with {\it tail}. We now evaluate three potential designs: no abstraction layer (\nth{1} row, as in self-attention), abstraction layer's transformation for all KQV (\nth{2} row) and our design (\nth{3} row). Results show that transforming $\mathbf{v}$ has a detrimental effect to the model performance, as we conjectured. Furthermore, improving the features of $\mathbf{q}$ and $\mathbf{k}$ for compression using our abstraction layer is very beneficial and it improves Top1 accuracy by 1.9. \\ 

\begin{figure*}[t]
    \centering
    \includegraphics[width=0.98\textwidth]{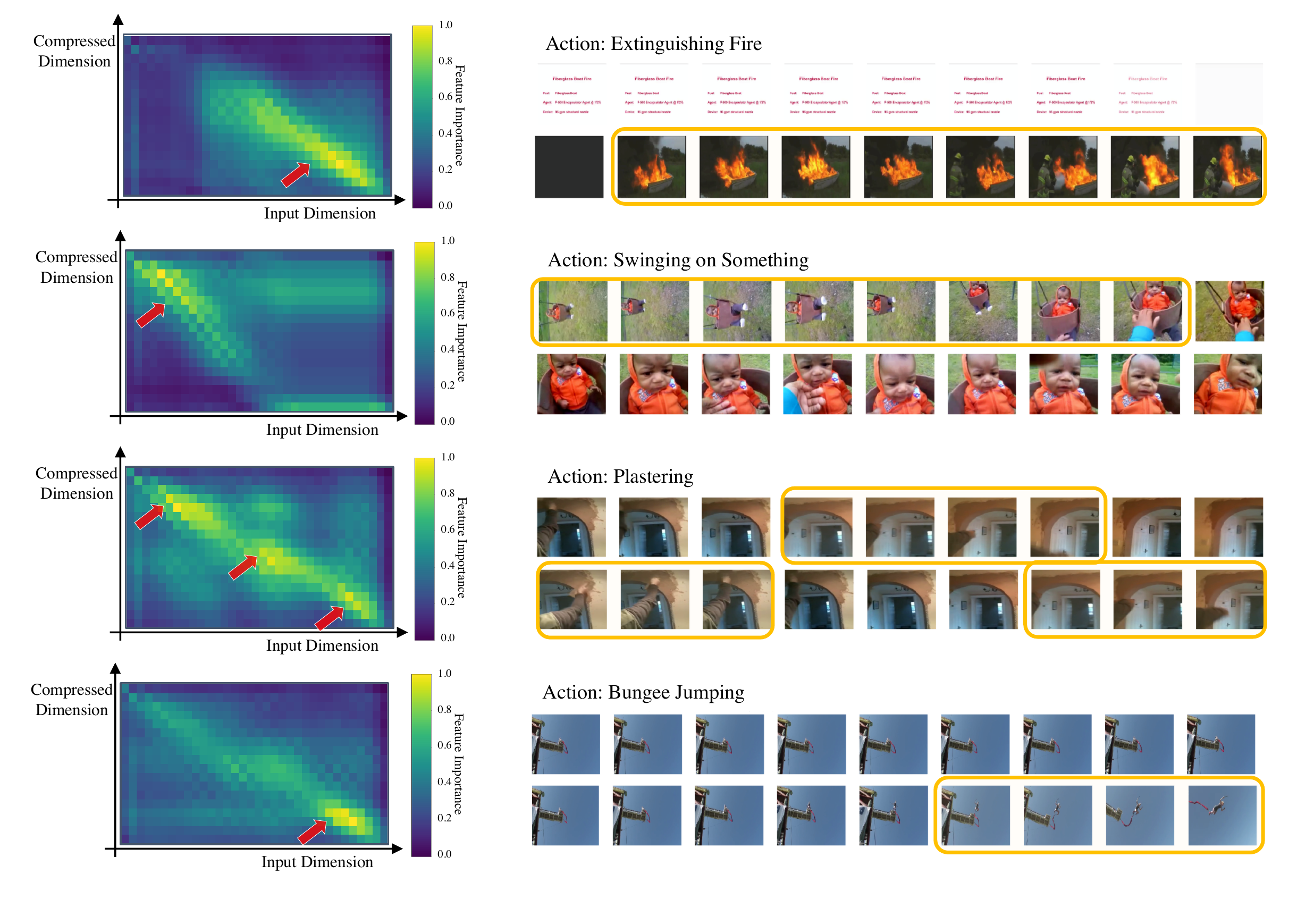}
    \vspace{-5mm}
    \caption{\it \small We show the temporal attention map $\mathbf{M}$ (eq.~\ref{eq:SFC}) and the corresponding input frames, for four examples. On all videos, SFC is able to attend to the frames relevant for action recognition. For example, only the the second half of the first video contains a fire, and only the first half of the second video contains a baby swinging.}
        \vspace{-2mm}
    \label{fig:visual}
\end{figure*}

\noindent {\it Convolution designs} (table~\ref{tab:aba-e}). We evaluate different linear transformations for $\mathbf{q}$ and $\mathbf{k}$ in SFC. Results show that employing 3D temporal kernels only in the temporal channel is the best option, which is coherent with the goal of SFC of  temporal compression. \\



\begin{table}[t]
    \small
        \centering
         \begin{tabular}{llccc} 
			\toprule
			Method & Input Size & Usage & FLOPS & Top1 \\
			\midrule
			Single Crop & 64$\times$1 & 100\% & 54G & 64.8 \\
			Uniform Sampling  & 64$\times$30 & 100\% & 1620G & 76.7 \\
			Dense Sampling  & 64$\times$180 & 100\% & 9720G & 78.2 \\
			\hline
			SFC, $\tau = 2$ & 256$\times$1 & 50\% & 247G &  77.4	\\
			SFC, $\tau = 4$ & 256$\times$1 &  25\% & 195G &  75.2	\\
			SFC, $\tau = 8$ & 256$\times$1 &  12.5\% & 167G &  68.2 \\
			\hline
			SFC, $\tau = 2$ & 1024$\times$1 & 50\% & 988G &  \textbf{78.5}	\\
			 SFC, $\tau = 4$ & 1024$\times$1 &  25\% & 780G &  77.2	\\
			SFC, $\tau = 8$ & 1024$\times$1 &  12.5\% & 668G &  74.0 \\
			\bottomrule
		\end{tabular}
    \vspace{-2mm}
    \caption{\it \small Results on ActivityNet v1.3 using  Slow I3D 8$\times$8.}
    \vspace{-3mm}
    \label{tab:anet}
\end{table}

\section{Extension of SFC to Untrimmed Videos}
\label{sec:res2}
Although SFC is proposed for trimmed videos, we now explore it in the context of untrimmed content. The main challenge with these videos is that their temporal length varies drastically, from a few seconds to potentially many hours. Our SFC model can however take as input around 4k frames (${\sim}2$ minutes of videos at 30FPS) within the current memory constraints. To overcome this limitation, we combine SFC with a uniform sampling strategy, which is the popular to-go solution for untrimmed videos, as dense sampling is unfeasible for long videos. 

We experiment on ActivityNet v1.3t~\cite{caba2015activitynet} using a Slow-Only I3D 8$\times$8 backbone. Since this dataset contains videos that are on average shorter than 10 minutes, we uniform sample 8 clips from each video. Then, we sample a certain number of consecutive frames from each clips and finally we concatenate all these and feed them into our network. To understand how the number of sampled frames affects the performance, we experiment with two input sizes: 256 and 1024 frames. For the 256 variant, we sample 32 frames from each of the 8 video clips, while for the 1024 variant we sample 128 frames (table~\ref{tab:anet}).  
For reference, we also report results using Single, Uniform and Dense sampling. Except for Single, the other two are substantially slower solutions compared to SFC. We also experiment with different compression rates ($\tau$) to evaluate how the performance drops as SFC compresses more and more aggressively. 

Results show that SFC can already improve over Uniform sampling performance using only 15\% of its FLOPS (76.7 US vs 77.4 SFC with $\tau=2$ and input of 256). SFC can also outperform Dense sampling while improving its FLOP by an order of magnitude (78.2 DS vs 78.5 SFC with $\tau=2$ and input of 1024). While SFC was designed to improve short clip classification performance, these results show that it can also be extended to longer videos. Importantly, note how SFC can be easily adapted to even longer videos than those in ActivityNet by increasing the number of uniformed sampled clips and reducing the number of frames sampled from each clip, as our results show that using 256 frames as input already achieves very competitive performance compared to sampling 1024 (77.4 vs 78.5).

\section{Visualization} \label{sec:vis}
 We show four examples in figure \ref{fig:visual} to better interpret the functionality of SFC module. In each example, we visualize the temporal association map $\mathbf{M}$ together with raw input frames. Yellow pixels mean a higher feature response that indicates an important region. On the right site, we cover the informative frames with transparent yellow boxes. From the first two rows, we can observe that when there are background or shot transitions, our SFC module is able to ignore this noise. Also, when the background is not changing, SFC is able to choose regions with clear motion change. In the example of plastering, the affinity map responds highly in the temporal region where a hand appears. In the example of bungee jumping, the affinity map responds highly on the temporal region where the actor starts to jump. This implies that SFC module can smartly select important feature.



\section{Conclusion} \label{sec:concl}
We presented Feature Compression Plugin (SFC), a flexible plugin that can be inserted into several existing action recognition networks to perform video level prediction in one single pass. Our compression design is guided by the idea that dropping non-informative feature can boost inference speed and not hurt model performance. By avoiding dense sampling, our method can achieve 6-7x speedup, 5-6x less memory usage and slightly higher Top1 accuracy, compared to the popular 30-crops dense sampling.

\clearpage
{\small
\bibliographystyle{ieee_fullname}
\bibliography{egbib}
}

\end{document}